\documentclass[accepted]{uai2024} 
                        
\usepackage[american]{babel}

\usepackage{natbib} 
\usepackage{mathtools} 
\usepackage{booktabs} 
\usepackage{tikz} 

\usepackage{times}  
\usepackage{helvet}  
\usepackage{courier}  
\usepackage{graphicx} 
\def\UrlFont{\rm}  
\usepackage{natbib}  
\usepackage{caption} 
\usepackage{algorithm}
\usepackage{algorithmic}
\usepackage{amsmath}
\usepackage{amssymb}
\usepackage{adjustbox}
\usepackage{caption}
\usepackage{subcaption}
\usepackage{xcolor}
\usepackage{amsfonts}
\usepackage{bm}
\usepackage{multirow}
\usepackage{amssymb}
\usepackage{newfloat}
\usepackage{listings}

\title{Structural Causality-based Generalizable Concept Discovery Models}

\author[1]{\href{mailto:<sanchit@virginia.edu>}{Sanchit Sinha}{}}
\author[1]{Guangzhi Xiong}
\author[1]{Aidong Zhang}

\affil[1]{%
    Computer Science Dept.\\
    University of Virginia\\
    Charlottesville, VA, USA
}

\begin{document}

\maketitle

\begin{abstract}
The rising need for explainable deep neural network architectures has utilized semantic concepts as explainable units. Several approaches utilizing disentangled representation learning estimate the generative factors and utilize them as concepts for explaining DNNs. However, even though the generative factors for a dataset remain fixed, concepts are not fixed entities and vary based on downstream tasks. In this paper, we propose a disentanglement mechanism utilizing a variational autoencoder  (VAE) for learning mutually independent generative factors for a given dataset and subsequently learning task-specific concepts using a structural causal model (SCM). Our method assumes generative factors and concepts to form a bipartite graph, with directed causal edges from generative factors to concepts. Experiments are conducted on datasets with known generative factors: D-sprites and Shapes3D. On specific downstream tasks, our proposed method successfully learns task-specific concepts which are explained well by the causal edges from the generative factors. Lastly, separate from current causal concept discovery methods, our methodology is generalizable to an arbitrary number of concepts and flexible to any downstream tasks. 
\end{abstract}

\section{Introduction}
Deep Neural Networks (DNNs) have successfully disrupted multiple avenues of daily life - ranging from language understanding to image generation  \citep{he2016deep,vaswani2017attention}. However, recent criticism of DNNs stems from their lack of \textit{explainability} wherein the inner workings of the model are completely shielded from the stakeholders and are used as a black box. This makes their usage in applications where focus on due and diligent process is paramount, such as banking, healthcare, etc., extremely limited negating the effects of revolutionary performance. Although post-hoc methods provide a rudimentary way of interpreting these models, the need of the hour remains designing interpretable architectures themselves. With larger architecture sizes becoming commonplace, abstract explainability approaches, which align with human thought are the need of the hour. A recent line of research which entails explaining DNNs using `concepts' has been proposed. Concepts are high-level independent abstract entities shared among multiple similar data points that align with human understanding of the task at hand. However, the concepts learned by the models implicitly may be entangled with each other and are not aligned with human-understandable semantics. Hence, special care must be taken in learning concepts.

Disentanglement representation learning process \citep{bengio2013representation,higgins2018beta,kim2018disentangling,chen2018isolating} estimates a lower-dimensional representation of a high-dimensional input distribution by learning a set of mutually independent \textit{generative factors} which capture distinct factors of variation in the data. As these methods capture the inherent generative process of the distribution itself, they are aligned well with human understanding and hence, are highly interpretable. Once trained, any sample can be visualized theoretically as a mathematical combination of these aforementioned \textit{generative factors}. A recent line of literature \citep{marconato2022glancenets} has attempted to align abstract concepts with the aforementioned \textit{generative factors} - hence providing both control and interpretability.

Even though disentanglement representation learning makes it possible for the model to learn independent generative factors from data, it is not practical in real applications since the human-understandable concepts used in real-world tasks are usually entangled or even casually related to each other. Thus, another approach has been proposed to understand disentanglement using the lens of causality \citep{yang2021causalvae}. The motivation behind this approach is the assumption that disentangled latent representations act as endogenous causal variables for the data generation process. Causal representation learning attempts to incorporate and enforce graphical structure among the learned latent representation and assumes that the data generation process can be fully explained by transitions along graph edges. 
The task at hand then becomes simply to estimate the causal graph mapping from latent representations (endogenous) to the observed data (exogenous).

\begin{figure}
    \centering
    \includegraphics[width=0.48\textwidth]{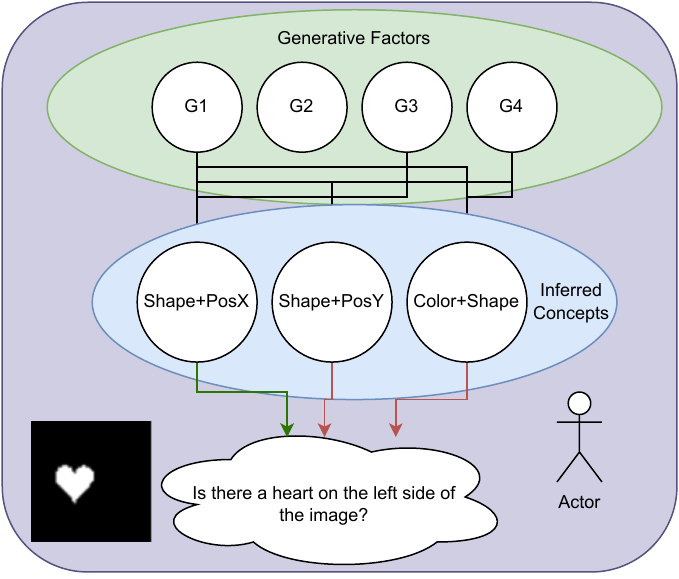}
    \caption{BOTTOM LEFT: Target image in question. Schematic figure to demonstrate why task-specific concepts are required. Even though the generative factors are known for the image, the concepts pertinent for correct classification can be causally related to multiple generative factors.}
    \label{fig:motivation}
\end{figure}

It is tempting to consider disentangled generative factors as concepts for explaining DNNs as generative factors are shared among all data points and are by definition independent. However, this line of thinking suffers from two distinct problems. The first problem relates to the fact that concepts used for explaining a specific DNN should be \textbf{task-specific}. For example, consider the example in Figure~\ref{fig:motivation} which attempts to infer understandable concepts for an image based on a particular ``task''. The image in question has been sampled from the d-Sprites \citep{dsprites17} dataset. The actor in question is interested in finding if the image contains a heart on the left of the center side of the image. Let us assume that the 1st concept encodes the shape and x-coordinate (position) of the image object. Now, we know that the shape and x-position are the only 2 factors that are responsible for making the task decision. Any other factors such as orientation, y-position, color, etc. are irrelevant to the task at hand. Now, let us assume that the task is switched to identifying if the heart is in the top half of the image. The generative factors remain the same, but the concepts change depending on the task, where now the concept encodes the shape and y-coordinate (position) of the image object. Hence, the goal is to encode relevant generative factors into concepts, not directly using generative factors as concepts.

The second problem relates to the \textbf{non-identifiability} of generative factors in a completely unsupervised setting. This problem is well studied in disentanglement literature \citep{locatello2019challenging,locatello2020disentangling}.  Let us assume in Figure~\ref{fig:motivation} that the generative factors are not known at all and need to be inferred from just the observed data and the task at hand. Concepts can encode multiple generative factors and are by definition abstract entities. For datasets with added complexity, the generative factors do not correspond to the actual meaningful and helpful cues, but a combination of their encoding in concepts makes for easy human understanding.

Hence, the problem of learning task-specific concepts requires a task-guided concept learning framework, incorporating disentanglement approaches for learning the generative factors. As concept learning is conditioned on the generative factors, they can be formulated using a Structural Causal Moel (SCM) \citep{pearl2009causality}. A SCM is a graph-based approach that estimates a set of endogenous variables (disentangled representations and concepts) with a pre-defined dependency structure based on a set of exogenous variables (task and inputs). Specifically, this paper:
\begin{itemize}
    \item Proposes a variational autoencoder-based model for learning disentangled generative factors, 
    \item Designs a structural causal model for mapping generative factors to task-specific concepts,
    \item Proposes a task structure to evaluate the flexibility and efficacy of the proposed framework, and  
    \item Utilizes datasets with known generative factors to evaluate the quality of generative factors encoded in concepts. 
\end{itemize}

\section{Related Work}

\subsection{Related work on Disentangled Representation Learning}

Disentanglement is a widely studied line of research and traces its origins from rudimentary wave analysis. In the Deep Learning era initial works such as
\citep{bengio2013representation,locatello2019challenging} attempted to learn the various \textit{factors of variations} of data and attempt to understand them using visualization and interpolation. A very popular disentanglement probabilistic model named - variational autoencoder (VAE) \citep{kingma2014auto} has been used with moderate success on small scale datasets to perform disentanglement. It constitutes an encoder and decoder which respectively map inputs to a regularized disentangled space and disentangled space to reconstructed inputs. Several improvements on the traditional VAE have been since proposed, for example \citep{higgins2018beta} - which propose $\beta$-VAE, a controlled disentanglement setup using an additional hyperparameter to amend the influence of the regularization term itself. Note that higher the regularization, the lower the quality of reconstruction. Usually a distribution similarity such as the Kullback–Leibler (KL) divergence is used for regularization where the similarity between the latent distribution and a standard Gaussian prior is minimized.  More variants such as FactorVAE \citep{kim2018disentangling} and $\beta$-TCVAE \citep{chen2018isolating} alter the basic regularization term and propose to directly penalize the total correlation - a practice which gives better optimization performance.

Even though the focus of seminal works like \citep{higgins2018beta,kim2018disentangling,chen2018isolating} was on unsupervised disentanglement, \cite{locatello2019challenging} proved that using unsupervised approaches are provably unidentifiable. In other words, it is impossible to learn \textbf{aligned} disentangled latent variables without any supervision. As a consequence, studies tried to improve identifiability using different kinds of supervision or biases \citep{locatello2020disentangling,shu2020weakly,shen2022weakly}.

\begin{figure*}[t]
    \centering
    \includegraphics[width=0.6\textwidth]{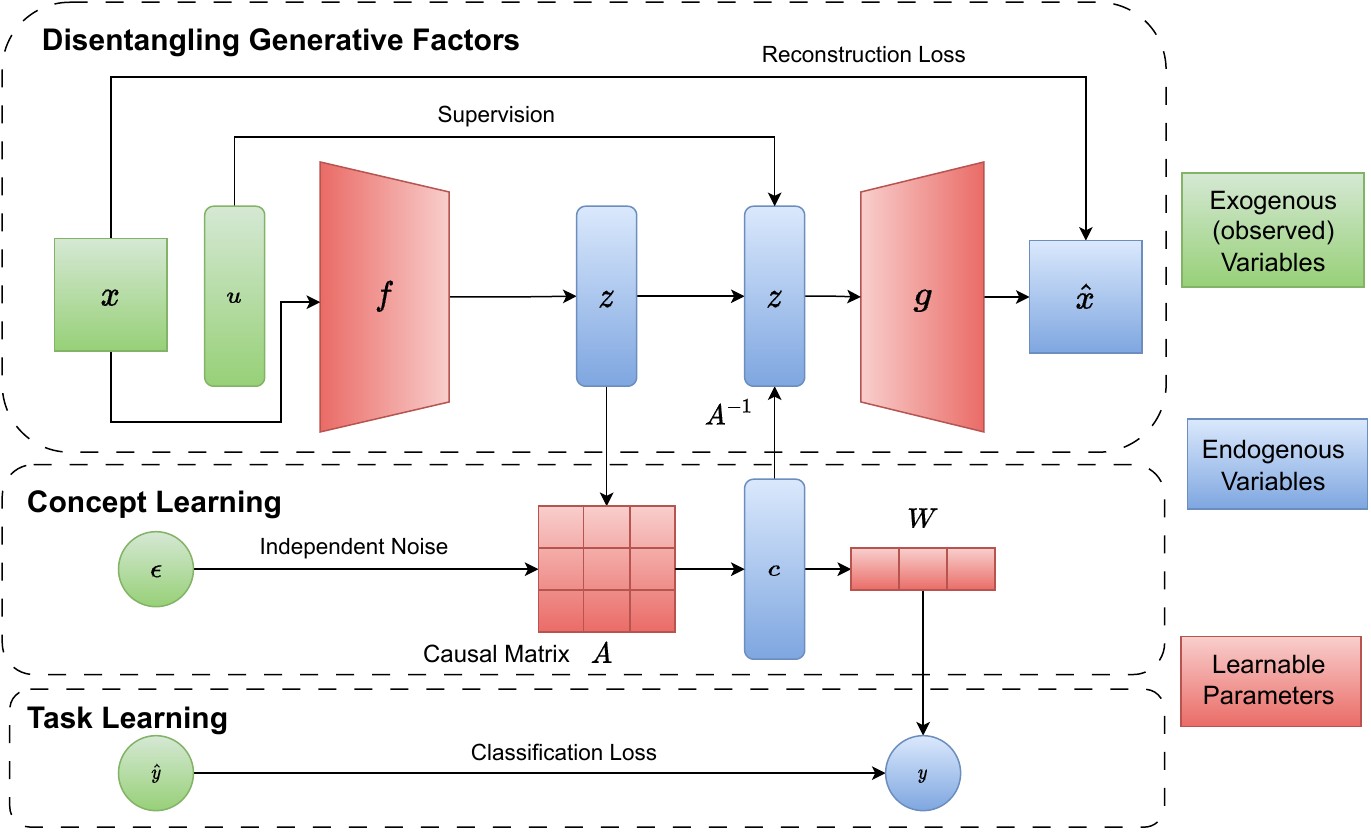}
    \caption{Schematic diagram of the proposed approach to learn task-specific concepts from disentangled independent representations (generative factors). Our proposed approach utilizes a variational autoencoder to learn representations of mutually independent generative factors ($\mathbf{z}$). The generative factors are transformed into concepts $\mathbf{c}$ by the Structural Causal Matrix $\mathbf{A}$. Finally, task prediction is performed by the weighted sum of the concepts $c$ and Weight Matrix $W$. $\hat{x}$ represents the reconstructed $x$. The exact formulation is detailed in Appendix. \textit{Green} blocks represent exogenous variables, i.e., variables that can be observed and are provided as input to the model. \textit{Blue} blocks represent endogenous variables, i.e., variables learned from the model. \textit{Red} blocks represent learnable parameters of the model. (Best viewed in color)}
    \label{fig:schematic-figure}
    \vspace{-4pt}
\end{figure*}

\subsection{Related work on Causality in Deep Learning and Structural Causal Models}

There are a series of works that incorporate structural causal models (SCMs) into the learning of generative models. CausalGAN \citep{kocaoglu2018causalgan} and CAN \citep{moraffah2020causal} are uni-directional generative models built on generative adversarial networks (GANs) \citep{goodfellow2014generative}, which assign SCMs to the conditional attributes while the latent variables are independent Gaussian noises. Since the ground-truth factors are directly fed into the generators in these two models as conditional attributes, they are not able to learn disentangled latent variables that are aligned with the factors. CausalVAE \citep{yang2021causalvae} and GraphVAE \citep{he2019variational} are two VAE-based models that involve SCMs during training. While CausalVAE directly assigns an SCM to the latent variables and adopts a conditional prior on them, GraphVAE imposes an SCM into the latent space of VAE by generalizing the chain-structured latent space proposed in LadderVAE \citep{sonderby2016ladder}, which is designed to improve expressive capacity of VAE.
\vspace{-2pt}

\subsection{Related work on Concept Based Explanations}
Explaining using abstract concepts in a post-hoc manner was first proposed in \citep{kim2018interpretability} by measuring the effect of low-level pre-selected concepts. Parallely, research has focused on aligning these pre-selected concepts during the model training process itself \citep{koh2020concept} - thus designing an interpretable architecture. Subsequent works have tried to remove the expensive concept curation step \citep{pedapati2020learning,jeyakumar2020can,heskes2020causal,o2020generative, yeh2020completeness,wu2020towards,goyal2019explaining,sinha2023understanding} by inferring concepts during training itself. The fundamental concept-based model maps input samples to a concept space and subsequently to the predictions. 

\noindent \textbf{Comparision to related work.}
The closest approaches to our work are GlanceNets \citep{marconato2022glancenets, sinha2024colidr} and CLAP \citep{taeb2022provable}. The former approaches utilize a variational inference backbone and attempts to align pre-defined concepts to the generative factors. This approach is hence, not comparable to ours as concepts need to be pre-defined for a task. The latter, CLAP is perhaps the closest approach to our work in intuition. However, CLAP incorporates the assumption that the generative factors are conditioned on the task itself. This is significantly different from our work where we condition the \textit{concepts} on the task instead of generative factors. Our approach is the first work to formulate concept learning as a causal learning mechanism from invariant generative factors.
\section{Methodology}
In this section, we first introduce the formulation of the structural causal model (SCM), which is designed to learn task-specific concepts. Next, we 
provide the procedure to learn disentangled representations using a variational autoencoder with the incorporation of our proposed SCM. Finally, we discuss the end-to-end training procedure followed by the final optimization objective. The schematic overview of our approach is presented in Figure~\ref{fig:schematic-figure}.  


\subsection{Learning Concepts using Structural Causal Model}

\subsubsection{Transforming endogenous latent representations}
Assume a set $Z= \{z_j\}_{j=1}^m$ consists of a mutually independent disentangled latent representation of generative factors of some data distribution, which is represented by a $m$-dimensional vector $\mathbf{z}$. Let ${C=\{c_i\}_{i=1}^n}$ denote a set of concepts describing some downstream task, represented by a vector $\mathbf{c}$ of size $n$. We propose to learn a causal representation of these task-specific concepts $\mathbf{c}$ based on the parent $\mathbf{z}$. Mathematically, we learn a structural causal model with $\mathbf{z}$ as independent parent nodes while concepts $\mathbf{c}$ are child nodes. The sets $Z$ and $C$ are ensured to be independent of each other and hence form a bipartite graph with directed edges from the set $Z$ to $C$. Mathematically we represent this procedure by 
\begin{equation}\label{eq:Z-to-C}
    \mathbf{c} = {h}(\mathbf{A}^{T}\mathbf{z}) + \mathbf{\epsilon}, \mathbf{\epsilon} \sim  \mathcal{N}(0,\,\mathbf{I}).
\end{equation}
In the equation, $\mathbf{A}$ denotes the causal matrix between the generative factors and concepts to be learned and $h$ corresponds to non-linear functions mapping transformed generative factors to concepts. $\mathbf{\epsilon}$ denotes the external independent noise multi-dimensional random variable in the process of data generation. The set of generative factors $Z$ and concepts $C$ form a bipartite graph $\Gamma$, implying no causal edges within $C$ or $Z$. In practice, we implement the transformation from $Z$ to $C$ as a linear layer. 
Specifically, each learnable concept $c_i$ is modeled as 

\begin{equation}
    \label{eq:z-to-c}
    {c_i} =  \mathbf{h}_i (\mathbf{A}_i\circ \mathbf{z};\eta_i) + \epsilon_i
\end{equation}
where $\mathbf{A}_i$ is a $m$-dimensional vector in $\mathbf{A} = \left[\mathbf{A}_1|\cdots|\mathbf{A}_n\right]$. The symbol $\circ$ denotes the element-wise multiplication. $\mathbf{h}_i$ is the non-linear mapping from the generative factors to the concept $c_i$, which is part of $h$. $\eta_i$ represents the parameters of the mapping $\mathbf{h}_i$. $\epsilon_i$ is a noise sampled from the standard Gaussian distribution $\mathcal{N}(0,1)$. The matrix $\mathbf{A}$ forms the core structural causal model (SCM) and models the graph edges between $Z$ and $C$ along with the functions $h$.

\subsubsection{Learning task-specific concepts}
As postulated by \cite{koh2020concept}, concepts are independent and diverse entities that are utilized for downstream tasks. We propose to predict the downstream task label $y$ as a weighted sum of the concepts with bias. 
\begin{equation}
    y = \sigma(w_0 + \sum^n_{i=1} w_ic_i) ~or~   \mathbf{Y} = \mathbf{W^T}\mathbf{C}
\end{equation}
where $\{w_i\}_{i=1}^n$ are the estimated weights and $w_0$ is the bias. The function $\sigma$ is the logistic function that transforms the outputs into labels.

\subsection{Learning Generative Disentangled Representations}
\subsubsection{Formulation of VAE}
As the concepts $C$ are learned on the basis of the generative factors $Z$, an essential step of our proposed method is to learn mutually independent generative factors from the data. 
We learn a disentangled representation $\mathbf{z}$ using parameters $\phi$ of the encoder $f$ by estimating the posterior distribution $q_{\phi}(z|x)$. 
The decoder $g$ of the VAE attempts to learn the details of the generative process of distribution $X$ conditioned on latent representation $z$, which is denoted by $p_{\theta^*}(x|z)$ where $\theta^*$ is the true parameter for the data generative process. 
With $x$ as the observed data and $\mathbf{u}$ as the labels of the generative factors $\mathbf{z}$, the probability distribution is parameterized by the endogenous variables $\mathbf{c,z}$ and exogenous variables $\mathbf{x,u,\epsilon}$.
\begin{equation}
    \mathbf{p_{\theta}(x,z,c,\epsilon|u) = \mathbf{p_{\theta}}(x|z,c,\epsilon,u)p_{\theta}(\epsilon,z,c|u)}
\end{equation}

\subsubsection{Noisy priors}
Using the noisy inference strategy proposed in CausalVAE \citep{yang2021causalvae}, we can model the probability distribution $\mathbf{p_{\theta}(x|z,c,\epsilon,u)}$ with $\mathbf{p_{\theta}}(x|z,c)$ because the $\mathbf{\epsilon,u}$ do not affect the distribution of $\mathbf{x}$. 
It should be noted that estimating $\mathbf{p_{\theta}(x|z,c)}$ is equivalent to estimating $\mathbf{p_{\theta}(x|z)}$, since $\mathbf{c}$ can be modeled as a function of $\textbf{z}$ as mentioned in Equations \ref{eq:Z-to-C} and \ref{eq:z-to-c}. Hence
\begin{equation}
    \mathbf{p(c|x)} = \mathbf{p(h(z)|x)} \equiv \mathbf{p(z |x)} 
    \label{eq:c-independence}
\end{equation}

Mathematically, given an encoding function $f$ with parameters $\phi$ and a decoding function $g$ with parameters $\theta$, the reconstruction can be formulated on the latent representation $\mathbf{z}$:
\begin{equation}
    \mathbf{p_{\theta}(x|z,c) = p_{\xi}(x-g(z))}
\end{equation}
while the representations can be 
\begin{equation}
    \mathbf{q_{\phi}(z,c,\epsilon|x,u) = q(z,c|\epsilon)q_{\zeta}(\epsilon - f(x,u))} 
\end{equation}
The probability distributions of $\mathbf{p}$ and $\mathbf{q}$ are formulated using the encoder and decoder with differentiable functions $f$ and $g$, respectively, such that given the input image $x$ and associated labels $u$ for generative factors, the encoder and decoder formulations can be represented as follows.
\begin{equation}
    \mathbf{z} = \mathbf{f(\mathbf{x,u})}+\zeta ,~~ 
    \mathbf{x} = \mathbf{g(z)}+\xi
\end{equation}
where $\zeta$ and $\xi$ are noise variables sampled from distinct standard Gaussian distributions.


\begin{figure}
    \centering
    \includegraphics[width=0.32\textwidth]{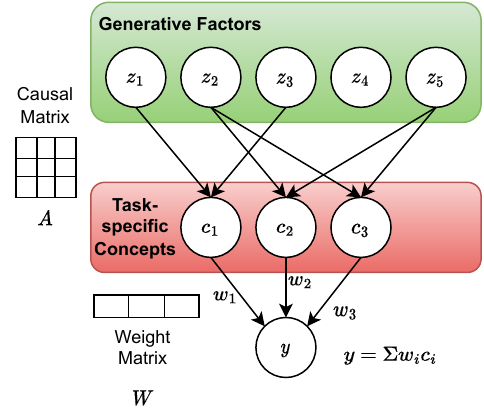}
    \caption{Visualized graphical structure for mapping generative factors to task-specific concepts. The generated concepts are child terminal nodes of parent nodes corresponding to independent generative factors. The sets $Z \in\{z_i \}^N_1$ and $C \in\{c_i \}^K_1$  form a bipartite graph $\Gamma(Z,C)$ with directed edges from $Z$ to $C$. The edges from $Z$ to $C$ are modelled by the Causal Matrix $A$ which captures the edge transitions.}
    \label{fig:concept-bipartite}
    \vskip -10pt
\end{figure}




\subsection{End-to-end Training}
\noindent \textbf{Estimating VAE parameters.} We utilize the Evidence Lower Bound (ELBO) criterion \citep{kingma2014auto} to estimate the probability distribution formulations in the VAE. Note that we have two sets of parameters in the encoder and decoder, parameterized by $\phi$ and $\theta$ respectively. We estimate the distribution $\mathbf{p_{\theta}(\epsilon,z,c|x,u)}$ by the corresponding ELBO formulations for VAEs \citep{kingma2014auto}.
\begin{align}
    \mathbb{E}_{q_\phi(\cdot|\mathbf{x})}[\log p_{\theta}(x|u)] \geq ELBO
\end{align}

The ELBO criterion can then be mathematically computed by ensuring conditioning on known distributions. Note that different from standard VAEs, we need to estimate both $\mathbf{z}$ and $\epsilon$ noise. Priors are introduced on $\epsilon$ using a pre-defined distribution. In our experiments, we utilize a standard normal distribution as prior for $\epsilon$, i.e., $p_{\epsilon}\sim \mathcal{N}(0,\mathbf{I})$. The ELBO criterion can be estimated as follows.
\begin{align*}
    \label{eq:elbo}
    ELBO = \mathbb{E}_{q} & [\mathbb{E}_{q_{\phi}(z,c|x,u)}[log ~p_{\theta}(x|z)]\\
    & - \beta_1\mathbb{D}_{KL}(q_{\phi}(\epsilon|x,u) || p_{\epsilon}(\epsilon)) \\
    & - \beta_2\mathbb{D}_{KL}(q_{\phi}(z,c|x,u) || p_{\theta}(z,c|u)) ]
\end{align*}
Note that $\mathbb{D}_{KL}(p|q)$ denotes the reverse KL-divergence between distributions $p$ and $q$. The strength of KL divergence in the ELBO criterion is controlled by 2 tunable hyperparameters - $\beta_1,$ and $\beta_2$.

\noindent \textbf{Supervision using generative factor labels $\mathbf{u}$.} To ensure the identifiability of disentangled latent variables, we provide supervision using generative factor labels. We minimize the effect of the transformation from $Z$ to $C$ by ensuring the inverse transform maps back to the identifiable labels.

\begin{equation}
    l_{u} = \mathbb{E}_{qx} \mathbf{ \|\sigma(u({A^{-1}c})) - \sigma(uz) \|^2_2 }  
\end{equation}

\subsubsection{Ensuring edge weight integrity}
The causal matrix $\mathbf{A}$ maps the generative factors $Z$ to the concepts ${C}$. Due to the nature of SCM, the matrix $\mathbf{A}$ corresponds to a Directed Acyclic Graph. However, it is possible that during the training procedure, degnerate edge-weights are learned, due to imbalanced sampling of observed data. Hence, we utilize a regularization and thresholding procedure to ensure edge weight integrity.

For the regularization procedure, we ensure that there are no degenerate concepts among $C$, i.e., one concept does not dominate the causal graph. We regularize the matrix using Frobenius Norm which ensures that the entire matrix $\mathbf{A}$ does not assign 0 weight edges to the associated SCM.
\begin{equation}
    \label{eq:diversity}
    l_{{diversity}} = \|\mathbf{A}\|_F
\end{equation}

The goal of thresholding is to control the effect of a single matrix element (edge weight) dominating the causal graph, we clip the maximum and minimum values in the matrix $\mathbf{A}$ to $\{1,-1\}$ after normalization.
\begin{equation}
    \mathbf{A} = clip(\mathbf{A}\circ 1/max(\mathbf{||A||}),-1,1)
\end{equation}

\noindent \textbf{Task learning and concept diversity.} In our proposed approach, we evaluate only on binary tasks, however, without loss of generalizability, we can represent classification loss as with $N$ number of classes as:
\begin{equation}
    l_{clf} = -\sum^N_{i=1}y_i\log\hat{y}_i 
\end{equation}
In addition, we ensure concept diversity as mentioned in Equation~\ref{eq:diversity}.
The complete training objective with tunable hyperparameters $\alpha, \beta_1, \beta_2, \delta, \gamma$ is as follows. 
\begin{equation}
    \mathcal{L} = -ELBO (\beta_1,\beta_2) + \alpha l_u + \delta l_{clf} + \gamma l_{diversity}
    \label{eq:end2end}
\end{equation}
Note that $\beta_1$ and $\beta_2$ are hyperparameters controlling the effect of the KL divergence.

\section{Experiments}
\subsection{Dataset Descriptions}
\begin{itemize}    
\item \textbf{D-Sprites} \citep{dsprites17}. D-Sprites consists of procedurally generated samples from 6 independent generative factors. 
Each object in the dataset is generated based on two categorical factors (shape, color) and four numerical factors (X position, Y position, orientation, scale).
The 6 factors are independent of each other. The dataset consists of 737280 images. 
\item \textbf{Shapes3D} \citep{3dshapes18}. Shapes3D dataset consists of synthetically generated samples from 6 independent generative factors consisting of color (hue) of floor, wall, object (float values) and scale, shape, and orientation in space (integer values). The dataset consists of 480000 images. 

\end{itemize}
Refer Appendix for more details.

\subsection{Task Descriptions}
We construct downstream tasks using combinations of generative factors. For each task, we consider $n\in\{2,3\}$ generative factors on random. A task is defined as having the label ``$1$'' when all factors satisfy a pre-defined criterion and having the label ``0'' in other cases. For categorical factors, we consider the presence of the exact values as boolean Truth, while for continuous factors we use a threshold to decide the boolean Truth value. For instance, for the d-Sprites dataset task in Figure~\ref{fig:motivation}, we adjudge the label 1 when `X-position' is less than $0.5$ and the `shape' value is exactly the categorical value ``heart''. We ignore tasks that impute a very small training set. The number and description of the generative factors is given in Table~\ref{tab:task-desc} along with the number of tasks considered in our experiments. Further details on task construction can be found in the Appendix.

\begin{table}[h]
\resizebox{\columnwidth}{!}{%
\begin{tabular}{cccc}
\hline
Dataset   & \# of Factors & Description & \# of Tasks \\
\hline
\multirow{3}{*}{D-Sprites} &         \multirow{3}{*}{6}           &  Shape, Color,        &     \multirow{3}{*}{9}        \\
&                  & PosX, PosY, &  \\  
&                  &  Orient, Scale &  \\  
\hline
\multirow{2}{*}{Shapes 3D} &         \multirow{2}{*}{6}          &     floor, wall, object hue, & \multirow{2}{*}{12} \\
 &                   & scale, shape, orientation      &            \\
\hline
\end{tabular}}
\caption{Dataset and generative factors description and the associated task statistics.}
\label{tab:task-desc}
\vskip -10pt
\end{table}

\subsection{Architecture Overview}
\noindent \textbf{Variational autoencoder}. For the d-Sprites dataset, we utilize an encoder architecture consisting of 3 fully connected linear layers $[900,600,300]$. The input images are scaled to (96, 96). The activation function utilized is the Exponential Linear Unit (ELU). The latent dimension ($z_{dim}$) is kept as 16. The corresponding decoder consists of 4 fully connected linear layers [300, 300, 1024, 96*96]. 

For the Shapes3D dataset, the input images also contain color information. We utilize a convolutional encoder with 6 layers with [3, 32, 64, 64, 64, 16] channels each and $z_{dim}$ = 16. The decoder consists of 3 transpose convolutional layers [64, 64, 3]. The activation function used is ReLU.

\noindent \textbf{Discussions on choices around number of concepts}.
The number of generative factors $m$ is already known. For all our experiments, we keep the number of concepts the same as generative factors, i.e., $n=m$. Hence the shape of the transformation matrix $\mathbf{A}$ is $(n,n)$. We initialize the weight matrix as a unit vector $\mathbf{W} = \mathbf{1}_n$.
In our work, we consider the number of concepts to be equal to the number of generative factors (GF). This choice is made in accordance with the following assumptions:
\begin{itemize}
    \item The datasets used are procedurally generated, hence they would be \textbf{completely explained} by the number of generative factors. Hence, the number of concepts is upper bounded by the number of GF, i.e. $n$ $\leq$ $m$ where $n$ are number of concepts while $m$ are number of GF.
    \item For datasets with an unknown number of generative factors, human-understandable concepts are always abstract in nature and are regularly composed of multiple aggregated disentangled probability distributions. Hence, we can assume with that $n$ $\leq$ $m$ holds.
    \item For experiments, $n$ $ = $ $m$ has been utilized to maintain the identifiability of the disentangled GFs and the concepts (Equation~10). It is statistically easier to compute the inverse mapping from concepts to GFs by assuming $m = n$. However, we point out that this is not a required condition and $m < n$ is a perfectly valid assumption.
\end{itemize}

\subsection{Hyperparameter Settings}
The proposed framework is sensitive to two sets of hyperparameters, which control the VAE reconstruction and the concept learning, respectively. For d-Sprites, we utilize $\beta_1$ and $\beta_2$ as 1.0 and 0.6, while for the Shapes3D dataset, $\beta_1$ and $\beta_2$ are set as 0.8 and 0.8. The values of $\alpha$ and $\gamma$ are set to 0.5 for both datasets and $\delta$ is set as 0.5 and 0.8 respectively for d-Sprites and Shapes3D. In our experiments, we observe that Shapes3D is much more sensitive to hyperparameters. One possible cause for this behavior is the encoding of colors in concepts making them unstable.
The learning rate for the VAE is set as 1e-3 with a linear warmup. The number of epochs used to train the end-to-end system is 50. However, in the first 10 epochs, the matrix $\mathbf{A}$ and $\mathbf{W}$ are not tuned, as we intend to ensure that VAE learns to disentangle the generative factors first and then guide concept learning.

\subsection{Evaluation}
Our evaluation framework evaluates the proposed method using its most important objectives. The methodology of evaluation is detailed as follows:
\begin{itemize}
    \item \textbf{Disentangled latent representation learning performance:} The most important component of the proposed method is learning independent generative factors. Most disentanglement methodologies utilize the Maximal Information Coefficient (MIC) for estimating the degree of information relevance between the known (true) distribution and the learned distribution.
    \item \textbf{Task performance: } As we only consider binary downstream tasks for estimating concepts, we utilize accuracy as metric and evaluate our approach on test set. 
    \item \textbf{Causal matrix learning:} The causal matrix learned is characterized by the relative weights of the transitions. Hence, we evaluate how accurately the relevant factors are identified (true positives). We also calculate the number of wrongly weighted causal edges from irrelevant generative factors (false positives) and the missed edges from relevant factors (false negatives). In our approach, we consider an entry ``False Positive'' if it is very close to the \textit{top-k}th lowest value.
    
\end{itemize}

\section{Results and Discussion}

\subsection{Quality of Latent Representation Learning}
\begin{figure}[h]
    \centering
    \includegraphics[width=0.42\textwidth]{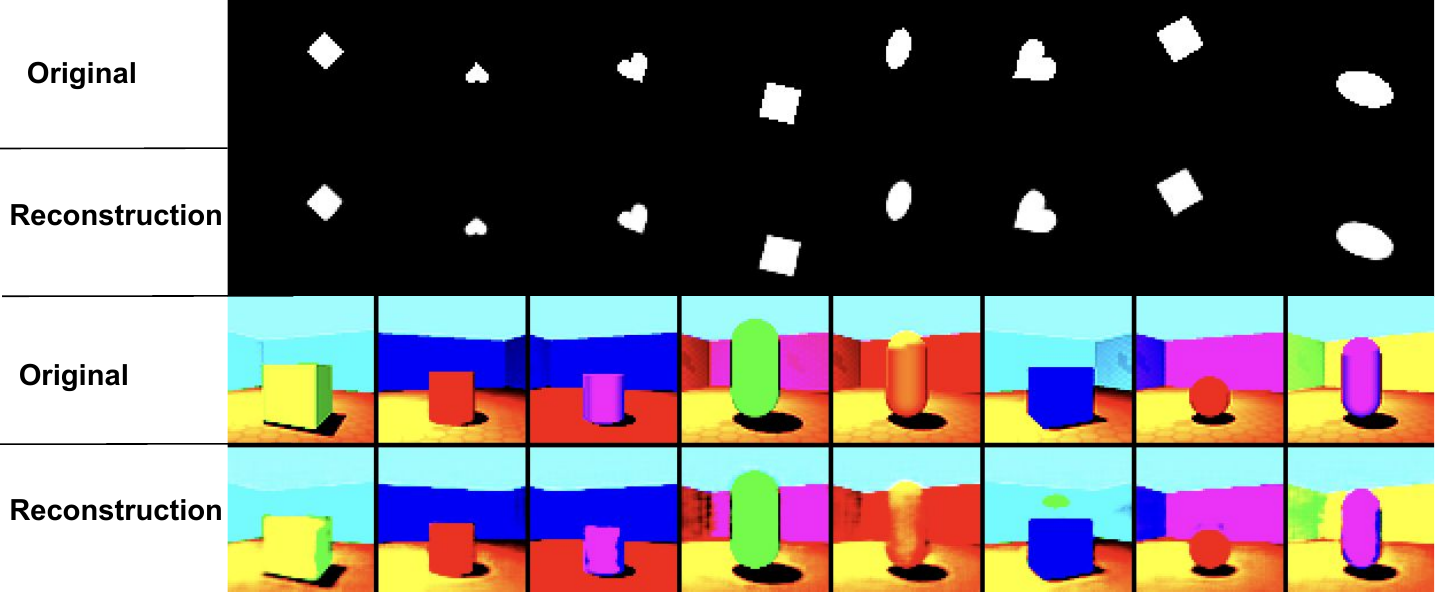}
    \caption{TOP: Original test set images and their associated reconstructions from the d-Sprites dataset. BOTTOM: Original test set images and their associated reconstructions from the Shapes3D dataset. As can be seen, the reconstruction quality is perceptibly similar. }
    \label{fig:reconstruction-vis}
\end{figure}

We start by visualizing the reconstructions to provide a qualitative analysis of the training of the VAE. Figure~\ref{fig:reconstruction-vis} shows the reconstructed images randomly sampled from the test sets of d-Sprites and Shapes3D. It can be observed that our approach can well reconstruct the input images with the learned latent representations. Next, we report the MIC scores for both datasets in Table~\ref{tab:mic}. In the first and second rows (Row 1 and Row 2), we compare the scores with a  $\beta$-VAE without and with noisy inference. Row 3 reports the MIC scores of our proposed approach. We observe that the MIC scores for noisy inference $\beta$-VAEs (Row 5) with supervision labels are the highest, followed by standard $\beta$-VAEs (Row 4) with supervision. Our method is between noisy $\beta$-VAE and supervised $\beta$-VAE. This is expected, as we attempt to learn both concepts and latent representations together which causes the disentanglement performance to drop slightly.
\vspace{-10pt}
    \begin{table}[h]
    \centering
    \begin{tabular}{c|c||c}
    \hline
          & \textbf{D-Sprites} & \textbf{Shapes3D}  \\
          \hline 
    $\beta$-VAE            &    0.48       &   0.41       \\
    Noisy $\beta$-VAE                           & 0.52          &  0.43         \\
    \textbf{Ours}                              & \textbf{0.77}          & \textbf{0.67}         \\
    Sup. $\beta$-VAE & 0.91          & 0.80         \\
    Sup. Noisy $\beta$-VAE                & 0.93          & 0.82       \\ 
    \hline
    \end{tabular}

\caption{MIC score values of our approach as compared to commonly utilized VAE formulations. Note that our approach learns concepts from disentangled latents.}
\label{tab:mic}
\vskip -10pt
\end{table}

\vspace{-4pt}
\subsection{Effect of Classification Loss Weight $\delta$ on Accuracy and Disentanglement Performance}
\begin{figure}
    \centering
    \includegraphics[width=0.36\textwidth]{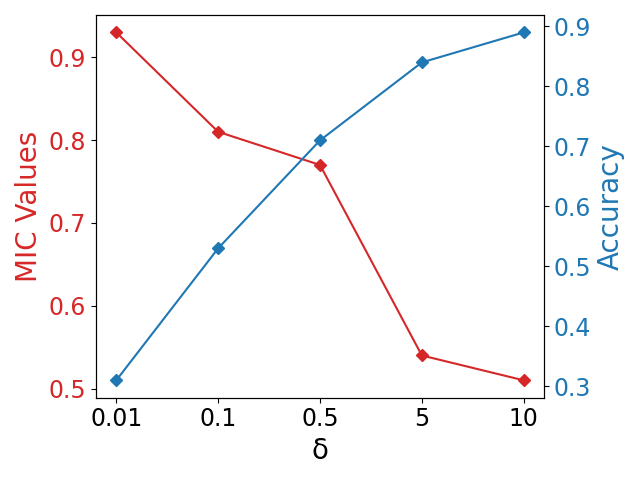}
    \caption{The effect of $\delta$ on overall MIC values and average task accuracy. We observe a tradeoff between task accuracy and MIC values with increasing values of $\delta$.}
    \label{fig:ablation-delta}
    \vskip -15pt
\end{figure}

We observe that the weight assigned to the classification loss is an extremely important hyperparameter. In Figure~\ref{fig:ablation-delta}, we show that a low $\delta$ results in higher MIC scores, but worse accuracy. On the other hand, with high $\delta$, we achieve very high accuracy but the MIC scores plummet implying a sensitive trade-off between task accuracy and disentanglement performance.

\begin{table}[h]
\begin{tabular}{c|cccc}
\hline
\textbf{D-Sprites} & \textbf{2-GF} & \textbf{3-GF} & \textbf{FP} & \textbf{FN} \\
\hline

Unconstrained         & 0.77             & 0.71             & 0.41                     & 0.26                     \\
w/ Thresholding        & 0.80             & 0.75             & 0.23                     & 0.17                     \\
\textbf{w/ Regularization}         & \textbf{0.93}             & \textbf{0.88}             & \textbf{0.13}                     & \textbf{0.09}                     \\
\hline
\textbf{Ground Truth}         &    0.99          &   0.97           &     0.01                &          0.02            \\

\hline

\end{tabular}
\\ 
\\
\\
\begin{tabular}{c|cccc}
\hline
\textbf{Shapes3D} & \textbf{2-GF} & \textbf{3-GF} & \textbf{FP} & \textbf{FN} \\
\hline
Unconstrained       &        0.44      &    0.41          &                 0.31    &            0.57         \\
w/ Thresholding        &     0.56        &    0.51          &         0.22             &           0.46          \\
\textbf{w/ Regularization}         &     \textbf{0.74}         &   \textbf{0.69}           &              \textbf{0.19}        &                \textbf{0.28}      \\
\hline
\textbf{Ground Truth}         &     0.93        &    0.93         &               0.01      &          0.07       \\
\hline
\end{tabular}
\caption{Performance of inferring the relevant and irrelevant generative factors for the learned causal  matrix $\mathbf{A}$ on the d-Sprites and Shapes3D datasets. The bottom-most row corresponds to the setting where the causal matrix is initialized based on ``known'' factor values. The first row denotes unconstrained $\mathbf{A}$, while second and third rows progressively add thresholding (clipping) and Frobenius regularization implying increasing constraints on learning $\mathbf{A}$.
}
\label{tab:dset-acc}
\vskip -15pt
\end{table}

\begin{figure*}[t]
    \centering
    \begin{subfigure}{.48\textwidth}
    \includegraphics[width=\textwidth]{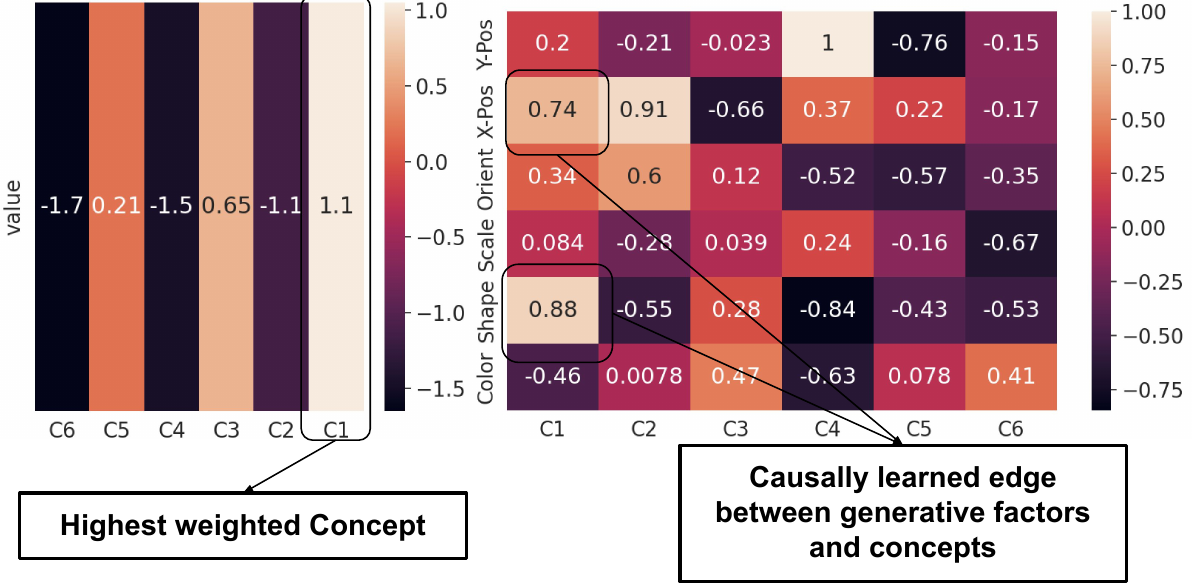}
    \caption{Learned task weights (left) and learned causal matrix $\mathbf{A}$\\ for the target binary task - ``heart shape \& left sided''. As can be \\ seen from the concept weights, concept ``C1'' causally encodes\\ information from ``shape'' and ``X-position'' and has the highest\\ edge weight during prediction.}
    \label{fig:dsprites-heatmap-t1}
    \end{subfigure}%
    \begin{subfigure}{.48\textwidth}
    \includegraphics[width=\textwidth]{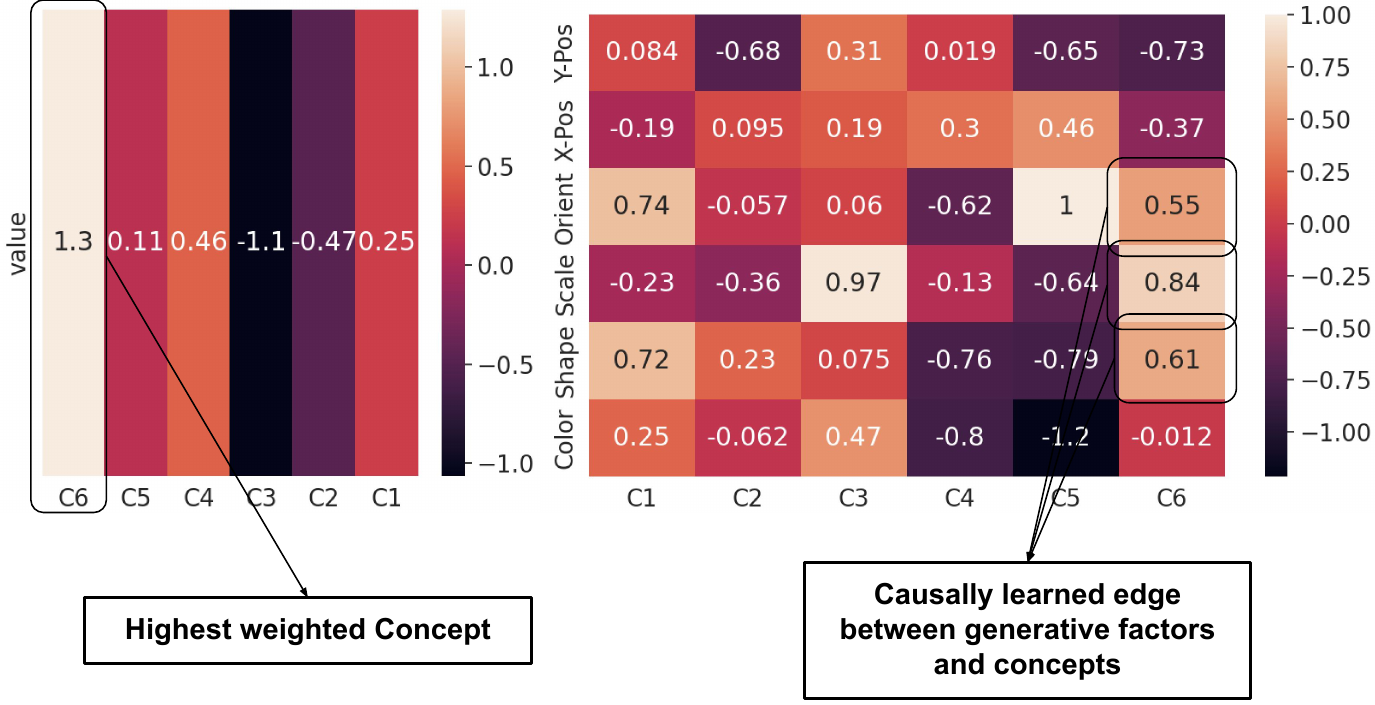}
    \subcaption{Learned task weights (left) and learned causal matrix $\mathbf{A}$ for the target binary task - ``Shape = heart \& position = top \& Scale $\geq 0.7$''. As can be seen from the concept weights, concept ``C6'' causally encodes information from ``Shape'', ``Scale'' and ``Y-position'' and has the highest edge weight during prediction.}
    \label{fig:dsprites-heatmap-t2}
    \end{subfigure}
    \caption{Visualizations of the learned edge weight matrices $\mathbf{W}$ and causal matrices $\mathbf{A}$ for 2 binary tasks for the d-Sprites dataset. Note that we plot the transpose of both matrices. The lighter the color, the higher the weight assigned to the edge is (Best viewed in color). Similar visualizations for the Shapes3D dataset can be found in Appendix.}
    \label{fig:dsprites-causal-analysis}
\end{figure*}

\subsection{Causal and weight matrix evaluation}
\noindent Inferring the causal matrix is tricky because multiple causal edges from the generative factors to the concepts can be learned. In addition, it is impossible to control exactly which concept encodes the appropriate generative factors for a task. Hence, we infer the matrix in reverse - we observe the index of the highest weight in the weight matrix $\mathbf{W}$ - the most relevant concept for task performance. Subsequently, we observe the highest matrix entries associated with this particular concept which corresponds to the edge weights from the generative factors. We demonstrate the process visually in Figure~\ref{fig:dsprites-heatmap-t1}.

Figure~\ref{fig:dsprites-heatmap-t1} represents the weight and causal matrix $\mathbf{W}$ and $\mathbf{A}$ for the binary task constructed using ``Shape == Heart \& left sided == True''. We observe that the concept with the highest weight is ``C1'' and the causally associated generative factors ``shape'' and ``x-position'' are properly learned, represented by the highest relative entries along the ``C1'' concept column. Similarly, for Figure~\ref{fig:dsprites-heatmap-t2}, highest weighted concept ``C6'' encodes the 3 generative factors - ``shape'', ``scale'' and ``y-position''. We calculate the average correctly learned relevant generative factors for all tasks, non-learned relevant factors (false negatives-FN) as well as the wrongly learned irrelevant factors (false positives-FP) in Table~\ref{tab:dset-acc} for D-Sprites and Shapes3D datasets respectively. In the table, 2-GF and 3-GF represent tasks with 2 and 3 relevant generative factors (GFs) respectively. We also report the performances comparing the effect of thresholding and regularization (Rows 2 and 3). As expected, thresholding and regularization improve the accuracy of learning relevant and removing non-relevant generative factors. We also compare our performance to ground truth intialization where we initialize the causal matrix $\mathbf{A}$ with relevant generative factors and do not learn the causal matrix (Row 4).

\section{Conclusion}
In this paper, we propose a novel methodology for causal concept discovery using disentangled representations of underlying generative factors. Our method is successful in identifying the causal relationships between the generative factors and the concepts specific to downstream tasks. Experiments performed on carefully curated downstream classification tasks sampled from datasets where the generative factors are well-defined demonstrate that our proposed method discovers correct causal relationships between generative factors and concepts. Our work opens doors for future research which unifies disentanglement and causal discovery of concepts.

\bibliography{uai2024-template}
\newpage

\onecolumn

\section{Appendix}
The Appendix is organized as follows:
\begin{itemize}
    \item Mathematical formulation of $\beta$-VAE
    \item Maximal Information Coefficient (MIC) calculations
    \item Detailed Task Descriptions
    \item Visual Reconstruction examples
    \item Additional Results: D-Sprites and Shapes3D
\end{itemize}

\section*{Mathematical formulation of $\beta$-VAE}
We show the exact mathematical formulation of the Expected Lower Bound (ELBO) criterion and the training objective for the Variational Autoencoder.
    Recall that we denote the encoder and decoder by $f$ and $g$ respectively. For a given input $\mathbf{x}$ and associated generative factor labels $\mathbf{u}$, the encoder parameterized by $\mathbf{x,u}$ learns the latent space representation $\mathbf{z}$ as follows:
    \begin{equation}
    \mathbf{z} = \mathbf{f(\mathbf{x,u})}+\zeta 
    \end{equation}
    The variable $\zeta$ is utilized for adjusting additional noise in the setup. Note that the encoder is deterministic when $\zeta$ is infinitesimally small. For increasingly complex data, sampling procedure for $\zeta$ plays a crucial role in determining the quality of representations and preventing adulteration with extra noise.
    Similarly, the decoder $g$, parameterized by the latent representations $\mathbf{z}$ is represented as follows:
    \begin{equation}
        \mathbf{\hat{x}} = \mathbf{g(z)}+\xi    
    \end{equation}
    The variable $\xi$ is utilized to adjust for any additional noise during the decoding phase. Note that the decoder is deterministic when $\xi$ is infinitesimally small. Similar to $\zeta$, $\xi$ plays a crucial role in determining the quality of reconstructions.

    The ELBO criterion is formulated as follows

    \begin{align*}
    ELBO = \mathbb{E}_{q} & [\mathbb{E}_{q_{\phi}(z,c|x,u)}[log ~p_{\theta}(x|z)]\\
    & - \beta_1\mathbb{D}_{KL}(q_{\phi}(\epsilon|x,u) || p_{\epsilon}(\epsilon)) \\
    & - \beta_2\mathbb{D}_{KL}(q_{\phi}(z,c|x,u) || p_{\theta}(z,c|u)) ] \label{eq:elboappendix}
    \end{align*}

    Note that the first term from Equation only depends on the input $\mathbf{x}$ and the latent representations $\mathbf{z}$. The quantity $\mathbb{E}_{q_{\phi}(z,c|x,u)}$ can be simplified into $\mathbb{E}_{q_{\phi}(z|x,u)}$ as per Equation~\ref{eq:c-independence} (main text) and can be easily calculated. The rest of the KL terms weighted by $\beta_1$ and $\beta_2$ are conditioned on known distributions and are tractable. For more details refer to \cite{yang2021causalvae}.

    Apart from the ELBO criterion, we also ensure that the inputs $\mathbf{x}$ and their reconstructions $\mathbf{\hat{x}}$ are mutually agreeable. The reconstructions $\mathbf{\hat{x}}$ are given as follows:
    \begin{equation}
        \hat{\mathbf{x}} = g(f(\mathbf{x})+\zeta)+\xi
    \end{equation}
    The training objective minimizes the reconstruction loss $\mathcal{L}_{rec}$ as follows:
    \begin{equation}
        \mathcal{L}_{rec} = l ( \mathbf{\hat{x}},\mathbf{x} )
    \end{equation}
    where $l$ is any symmetric loss. In our experiments, we utilize the Mean Square Error Loss. Apart from ELBO, reconstruction loss is also optimized in the training objective.

    Note that we experiment with both $\beta$-VAE and Noisy $\beta$-VAE (Table~\ref{tab:mic}). The only difference between the 2 approaches is the values of $\zeta$ and $\xi$, which are set to 0 for $\beta$-VAE. The End-to-end training criterion in the main text (Equation~\ref{eq:end2end}) also implicitly contains the Reconstruction loss $\mathcal{L}_{rec}$. It has been omitted to avoid confusion with other loss terms.

\section*{Maximal Information Coefficient calculations}
The Mutual Information Coefficient is proposed by \cite{kinney2014equitability} which can be calculated as an extension of Mutual Information. Sets $X$ and $Y$ are sampled from the learned latent distributions $\mathbf{z}$ and the latent labels $\mathbf{u}$ respectively. Mathematically, MIC can be calculated as follows.
\begin{equation}
    MIC[x,y] = max_{|X||Y|} \frac{I[X;Y]}{log_2(min(|X||Y|))}
\end{equation}
The quantity $I$ calculates the mutual information between the sets $|X|$ and $|Y|$. The quantity $log_2(min(|X||Y|)$ is dependent on the size of the sets.
As MIC is a measure of correlation, higher MIC values indicate a better-learned distribution $\mathbf{z}$. Note that a completely supervised VAE learning procedure would entail a MIC value of 1. On the other hand, a completely unsupervised approach ``learns'' the probability distribution, but has no way to actually assign a learned latent distribution to its ``correct'' label. This problem has been well documented in unsupervised disentanglement literature and is sometimes referred to as the Non-Identifiability problem of VAE.

MIC measures how well the learned representations correspond to the original distribution. As can be seen in Table~\ref{tab:mic}, the difference in the MIC values of the completely unsupervised VAEs and supervised VAEs is large. Our proposed method lies somewhere in the middle of both approaches.

\section*{Task Descriptions}
\subsection*{Split descriptions:} 
For each dataset, we utilize 70\% of the data for training the VAE and learning task-specific concepts and 30\% for testing. 
Due to the fact that every attribute value combination is present exactly once in the dataset - random splitting can make the sets imbalanced with respect to a particular attribute. We split the sets in 70/30 proportion of the shape, posX and posY attributes for the D-sprites dataset while considering floor, wall, object hue and shape for the Shapes3D dataset.
\subsection*{D-Sprites}
We create 9 distinct binary tasks for the D-Sprites dataset. The tasks are created by splitting the training set based on n $\in$ \{2,3\} factors. As the training set is randomly sampled, the number of training points in each task is different but is ensured to be label balanced. In Table~\ref{tab:tasks-dsprites} the first 6 rows represent tasks created using split conditions on 2 factors (2-GF) and the next 3 rows represent tasks created using split conditions on 3 factors (3-GF). The tasks are ensured to be label balance between positive and negative samples.

\begin{table}[h]
    \centering
    \begin{adjustbox}{max width=0.48\textwidth}
    \begin{tabular}{c|c}
    \hline
    Criterion    & Implication (Label=1) \\  
    \hline
    PosX $\leq$ 0.5 \& Shape==3    &  Left-sided Hearts\\
    PosX $\geq$ 0.5 \& Shape==2    &  Right-sided Ellipses\\
    PosY $\leq$ 0.5 \& Shape==1    &  Bottom Squares\\
    PosY $\geq$ 0.5 \& Shape==3    &  Top Hearts\\
    Scale $\leq$ 0.5 \& Orientation $\geq$ 3    &  Big right rotated\\
    Scale $\leq$ 0.5 \& orientation $\leq$ 3    &  Small left rotated\\
    \hline
    \multirow{2}{*}{PosY $\geq$ 0.5 \& Shape == 3 \& Scale $\geq$ 0.7 }  & \multirow{2}{*}{Top big Hearts}\\
    \\
    \multirow{2}{*}{PosX $\geq$ 0.5 \& Shape == 2 \& Scale $\leq$ 0.5 }  & \multirow{2}{*}{Left small Square}\\
    \\
    \multirow{2}{*}{PosX $\geq$ 0.5 \& Orientation$\geq$3 \& PosY $\geq$ 0.5 }  & \multirow{2}{*}{Top right rotated}\\
    \\
    \hline
    \end{tabular}
    \end{adjustbox}
    \caption{Binary Tasks for D-Sprites. We create 6 tasks based on splitting 2 factors (2-GF) and 3 tasks based on splitting 3 factors (3-GF).}
    \label{tab:tasks-dsprites}
\end{table}

\subsection*{Shapes 3D}
We create 12 distinct binary tasks for the Shapes3D dataset. Similar to D-Sprites, the tasks are created by splitting the training set based on n $\in$ \{2,3\} factors. In Table~\ref{tab:tasks-shapes3d} the first 6 rows represent tasks created using split conditions on 2 factors (2-GF) and the next 6 rows represent tasks created using split conditions on 3 factors (3-GF). The tasks are ensured to be label balance between positive and negative samples.

\begin{table}[h]
    \centering
    \begin{adjustbox}{max width=0.48\textwidth}
    \begin{tabular}{c}
    \hline
    Criterion    \\  
    \hline
    Floor $\leq$ 0.5 \& Wall $\leq$ 0.5 \\
    Floor $\geq$ 0.5 \& Object $\leq$ 0.5    \\
    Wall $\geq$ 0.5 \& Object $\leq$ 0.5    \\
    Floor $\leq$ 0.5 \& Shape==1    \\
    Wall $\geq$ 0.5 \& Shape==2    \\
    Object $\leq$ 0.5 \& Scale $\geq$ 0.5    \\
    Scale $\leq$ 0.5 \& Shape==3   \\
    \hline
    Floor $\geq$ 0.5 \& Scale $\geq$ 0.5 \& Orientation==1   \\
    Object $\geq$ 0.5 \& Shape == 1 \&  Orientation==1   \\
    Floor $\geq$ 0.5 \& Wall $\geq$3 \&  Object $\geq$ 0.5  \\
    Floor $\leq$ 0.5 \& Wall $\leq$3 \&  Object $\leq$ 0.5  \\
    Floor $\leq$ 0.5 \& Wall $\geq$3 \&  Object $\leq$ 0.5  \\
    Floor $\geq$ 0.5 \& Scale $\geq$0.5 \& Shape == 2  \\
    \hline
    \end{tabular}
    \end{adjustbox}
    \caption{Binary Tasks for Shapes3D. We create 6 tasks based on splitting 2 factors (2-GF) and 6 tasks based on splitting 3 factors (3-GF).}
    \label{tab:tasks-shapes3d}
\end{table}

\section*{Visual Reconstruction Quality}
\begin{figure*}[h]
    \centering
    \includegraphics[width=\textwidth]{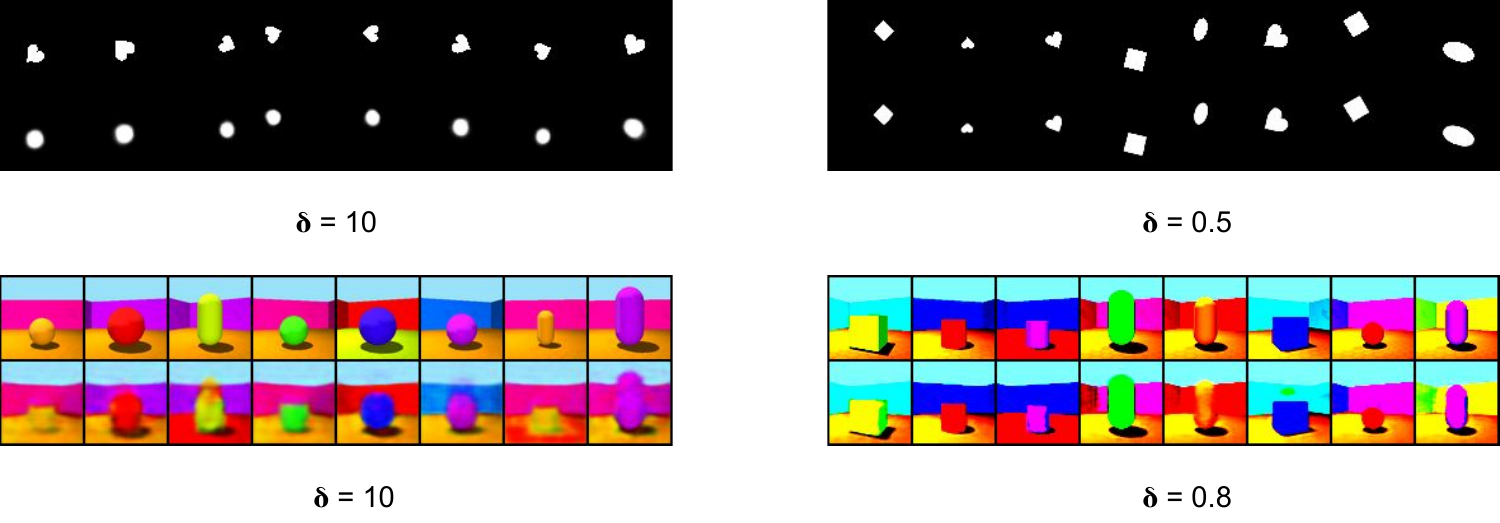}
    \caption{Varying $\delta$ values and associated quality of reconstruction. $\delta$ plays an important role in determining the tradeoff between task accuracy and reconstruction quality. The top rows of figures are original test samples and the bottom rows are reconstructed samples. We utilize $\delta$ values of 0.5 and 0.8 for D-sprites and Shapes3D respectively.}
    \label{fig:delta-hyp}
\end{figure*}

We report the visual reconstruction quality with varying values of hyperparameter $\delta$ controlling the strength of classification loss. In Figure~\ref{fig:delta-hyp} the top row shows the results of random original and reconstructed images from the D-Sprites dataset with 2 different $\delta$ values. As can be observed in the image on the left, the reconstructions lack sharp details. Similarly in the second row, The object colors are mixed on the left. Hence, an intermediate value of $\delta$ is required for the best accuracy and reconstruction performance tradeoff.

\section*{Additional results on D-Sprites and Shapes3D}
\begin{figure*}[h]
    \centering
    \includegraphics[width=\textwidth]{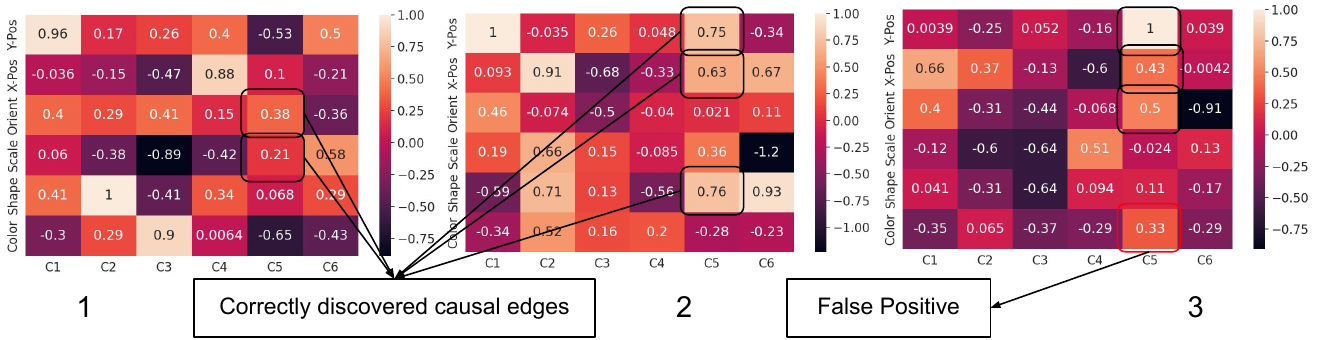}
    \caption{Examples of causal matrices for 3 sample tasks for D-Sprites Dataset. The factors are associated to the highest concept weight column. From Left-right: (1) 2-GF task: Scale and Orientation relevant generative factors (2) 3-GF task: Shape, Scale and PosX relevant factors and (3) 3-GF task: Orientation, PosX and PosY relevant factors, Color factor is false positive. We omit the weight matrix visualization.}
    \label{fig:example-d-sprite}
\end{figure*}

\begin{figure*}
    \centering
    \includegraphics[width=\textwidth]{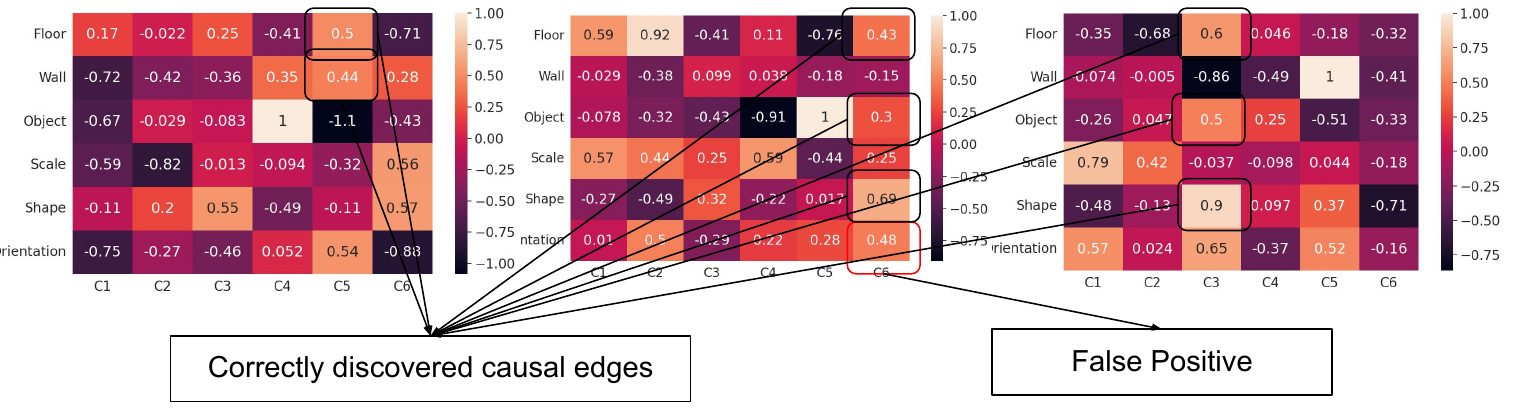}
    \caption{Examples of causal matrices for Binary tasks for Shapes3D Dataset.The factors are associated to the highest concept weight column. From Left-right: (1) 2-GF task: Floor and Wall Hue relevant generative factors (2) 3-GF task: Floor, Object and Shape relevant factors Orientation is false positive. and (3) 3-GF task: Floor, Object and Scale relevant factors. We omit the weight matrix visualization.}
    \label{fig:example-shapes3d}
\end{figure*}

Figures~\ref{fig:example-d-sprite} and Figure~\ref{fig:example-shapes3d} show the causal matrices for 3 distinct tasks for both datasets respectively. We also demonstrate some false positive edge weights inadvertently learned by our proposed method.

\end{document}